\documentclass[11pt]{article}
\usepackage{SOTA_style}

\usepackage{lipsum}  

\title{State of the Art: Image Hashing
}

\author{Rubel Biswas\\
    \href{mailto:rbis@unileon.es}{\texttt{rubel.biswas@unileon.es}} 
    \and Pablo Blanco-Medina\\
    \href{mailto:pblanm@unileon.es}{\texttt{pablo.blanco@unileon.es}}  
    }
\date{}

\begin{document}
{\setstretch{.8}
\maketitle
\begin{abstract}
Perceptual image hashing methods are often applied in various objectives, such as image retrieval, finding duplicate or near-duplicate images, and finding similar images from large-scale image content. The main challenge in image hashing techniques is robust feature extraction, which generates the same or similar hashes in images that are visually identical. 
In this article, we present a short review of the state-of-the-art traditional perceptual hashing and deep learning-based perceptual hashing methods, identifying the best approaches.
\noindent

\noindent

\end{abstract}
}



\section{Introduction}

The rise of the internet and smart devices, such as mobiles and digital cameras, has provided facilities to capture, store and share huge amounts of images and videos. Nowadays, the digital world experiences unauthorized distribution and illegal access of multimedia files by cybercriminals. Moreover, these types of crimes are carried out on computers or networks to spread illegal information, malware, images, copy attacks, or other materials. Indeed, cybercrime is one of the most effective ways to earn money in the criminal world. Besides, it is difficult to estimate correctly the actual cost of cybercrime. While the financial losses in business and effects in public security implications due to cybercrime can be significant.

\par 
Some specific types of cybercrimes are experiencing frequently by the internet and computer or smart device users, for instance, Credit card fraud; Phishing; Illegal Content; Identity theft; Software piracy, etc. Among them, sharing and distributing fake images, where fake/duplicate images could undergo various kinds of manipulation such as salient object-changing, or color-changing, are considered highly distressing and offensive. Because, nowadays, with the extensive use of low-cost and even free editing software, the cybercriminal can easily generate a fake/tampered image. With these editing tools, professional forgers generate multiple copies of an image with different digital representations such as rotation, compression, watermark embedding, editing, and tampering of an image, from the original one by keeping the actual visual contents. 

\par

Multimedia authentication and security are very demanding and challenging due to the advancement of digital forgery at a significant level. In order to protect such crimes and to support the Law Enforcement Agencies (LEAs), additional layers of prevention are required. Apart from social awareness, and cybersecurity strategy such as the internet or software security to securing systems, networks, and data, computer vision techniques such as perceptual image hashing \cite{ID2_15RobHashAuthentication}, can be applied in such domains of cybersecurity to detect, stop and respond to sophisticated crimes such as distributing fake images or illegal Tor domains classification on darknets~\cite{Nabki_17ClassIllegalTORText}. 

\par

To ensure the security of multimedia content, many researchers have presented similarity-based image detection or retrieval methods based on cryptographic hash functions, called perceptual hash functions or perceptual image hashing \cite{Yuenan12RobImaHashGabor, Tang2_16RobImaHasColor, Qin16PerImaHashSalientStrucFeat, Sajjad19RobImaHashSmartIndEnv, Qin19PercImaHashingLBP,Rubel2020F-DNS}. These methods are used in a wide variety of tasks, such as image retrieval, image authentication, digital watermarking, image copy detection, tamper detection, image indexing, multimedia forensics,
and reduced-reference image quality assessment \cite{yu2016iprivacy, du2020perceptual}. 

\par

Hashing methods extract certain features of an image to produce a $64$ or $128$ bits/numerical values, called hash code, based on these features. Perceptual hash functions have presented to form the “perceptual equality” of image content. 

\par

Recently, significant research has been done on deep hashing, which is the combination of perceptual hashing with deep learning techniques \cite{Liu2016Deep_supHash,ying2019locality,gu2019clustering,gu2019multi}. These deep hashing methods are applied for retrieving or detecting similar images from large scale datasets.

\par

In this document, we focus on the revision of the state-of-the-art traditional perceptual hashing \cite{Schneider1996RobustSignature} methods and deep learning-based hashing methods. Besides, we evaluate the robustness of four traditional perceptual hashing methods, Ring Partition and Invariant Vector Distance (RP-IVD) \cite{Tang1_16RobImaHashRingPartition}, Selective Sampling for Salient Structure Features (SS-Salient-SF) \cite{Qin16PerImaHashSalientStrucFeat}, pHash \cite{Zauner10ImplemPercImaHashFunc}, and F-DNS \cite{Rubel2020F-DNS}, using a state-of-the-art USC-SIPI dataset \cite{USC-SIPI}. Lastly, we present our conclusions in Section \ref{subsec:conclusions}.

\section{Perceptual hashing}\label{subsec:perceptual_hashing}

\subsection{Traditional perceptual hashing}
Perceptual hashing methods have traditionally been robust against certain types of attacks, like digital watermarking, noise addition, contrast adjustment or scaling, but not against rotations \cite{Fridrich_00RobHashWatermarking} or compression.
Several attempts have been made to decrease the impact of these issues \cite{Tang3_14RobImaHashDCT,Swaminathan06RobSecImaHash, Yuenan12RobImaHashGabor}. Nevertheless, enhanced performance under such circumstances usually resulted in increased sensitivity to other problems, such as contrast adjustment, salt and pepper noise \cite{Ahmed_10SecRobHashImaAuthentic}, tampered regions \cite{ID2_15RobHashAuthentication}, or watermark embedding \cite{Monga1_07RobSecImaHashNNMF}.

pHash \cite{Zauner10ImplemPercImaHashFunc} is a very well-known hashing approach in the literature. In this approach, firstly, the input image resized to $32\times32$ or $16\times16$ pixels and then applied DCT on it to obtain DCT coefficients. Later, the low-frequency DCT coefficients, omitting the lowest frequency coefficients, were selected for hash extraction. 

Davarzani et al. \cite{Davarzani_16PerImaHashLBP} presented the center-symmetrical local binary pattern (CSLBP) for representing the perceptual image content, proving to be useful for tampering detection. Tang et al. \cite{ZhenjunTang18ImageHashColor} constructed an image hash based on the angle of color vectors of color images, extracting the histogram from color angles and then compressing it to make a short hash. 

An image hashing system based on salient structure features proposed by Qin et al. \cite{Qin16PerImaHashSalientStrucFeat} which can be applied in image authentication and retrieval. In order to obtain the fixed length of the image hash, they conducted pre-processing in the input image first then salient edge detection was applied to extract a series of non-overlapping blocks containing the richest structural information according to the edge binary map. Later, Dominant DCT coefficients of the sampled blocks with their corresponding position information were retrieved as the robust features to compress to produce the final hash. 

Tang et al. \cite{Tang1_16RobImaHashRingPartition} incorporated ring partition and invariant vector distance to introduce an image hashing method for enhancing rotation robustness and discriminative capability. They mainly extracted the statistical features from image rings in perceptually uniform color space, i.e., CIE L*a*b* color space, and the Euclidean distance between vectors of these perceptual features were used to generate the image hash.

For person re-identification tasks, Fang et al. \cite{WenFang18PerHashPersonReID} characterized their images using perceptual hashing. They extracted low-level color and gradient features from an image, generating a hash feature map using the quantified features. After that, the histogram, mean vector, and co-occurrence matrix were extracted from the map central area to describe a person.

\subsection{Deep hashing} Due to the extensive growth of visual content on the Web, such as personal images and videos, retrieving/searching visually relevant or duplicate multimedia contents from very large databases are required. Besides, a database with tons of images is very common nowadays, and search through the database, especially linear search, would be costly in terms of time and memory. So recently, deep hashing, which is perceptual hashing based on deep learning architecture, is becoming an essential technique for large scale image retrieval \cite{Xia2014Supervised_hash,Liu2016Deep_supHash,jin2019unsupervised}.

\par

Jin et al. \cite{jin2019unsupervised} proposed a method that consisted of using semantic information from the Convolutional Neural Network (CNN) feature layer. Specifically, they used the VGG-19 \cite{Kang2018VGG19} model to preserve the information from feature space into hashing space, minimizing quantization loss between binary codes and hashing codes, increasing the information provided by each bit in the codes by using the highest information entropy. 

Gu et al. \cite{gu2019clustering} proposed an unsupervised end-to-end deep hashing framework for image retrieval, named Clustering-driven Unsupervised Deep Hashing (CUDH), consisting of training discriminative clusters recursively by a soft clustering model and generating binary codes with high similarity response. Subsequently, they also employed the aggregated clusters as an auxiliary distribution to generate hash codes. In another work \cite{gu2019multi}, Gu et al. presented a patch-based hashing framework for content-based medical image retrieval, applied to breast cancer diagnosis. They computed semantic similarity between random patches from low-magnification images by estimating the link propagation from the labeled high-magnification images. The hash codes of patches were learned by examining both local similarity and global labels.

\section{Discussion and Conclusions} 
\label{subsec:conclusions}

The authentication of multimedia contents is a vital issue in multimedia information security and related applications for protecting image integrity. This document presents a review of the state-of-the-art traditional perceptual hashing and deep perceptual hashing methods.
These algorithms are applied in different fields such as image retrieval, image authentication, digital watermarking, image copy detection, tamper detection, image indexing, and multimedia forensics \cite{Ahmed_10SecRobHashImaAuthentic,Swaminathan06RobSecImaHash,Ahmed_10SecRobHashImaAuthentic,ID2_15RobHashAuthentication,Monga1_07RobSecImaHashNNMF}.

In order to compute the similarity between hash codes, the correlation coefficient function is used in the literature \cite{ID7_14RobImaHashNMF}. Let $ H_1 = [h_1^{(1)},h_2^{(1)}, \ldots , h_l^{(1)}] $ and $ H_2 = [h_1^{(2)},h_2^{(2)},\ldots,h_l^{(2)}]$ be two image hashes where $l$ is the hash length and $h_i^{(\cdot)}$ is the $i$-th component of any of the hash codes. Thus, the correlation coefficient of $H_1$ and $H_2$ is calculated by means of Equation \eqref{eq:corrCoeff}:

\begin{equation}\label{eq:corrCoeff}
S(H_1,H_2)=\frac{\sum_{i=1}^{l}(h_i^{(1)}-\mu_1).(h_i^{(2)}-\mu_2)}{\sqrt{\sum_{i=1}^{l}(h_i^{(1)}-\mu_1)^2} \cdot \sqrt{\sum_{i=1}^{l}(h_i^{(2)}-\mu_2)^2} + \xi} ,
\end{equation}
where $ \mu_1 $ and $ \mu_2 $ are the mean values of $H_1$ and $H_2$, respectively. Moreover, $\xi$ is a small constant to avoid division by zero. The range of the correlation coefficient $S$ is $[-1, 1]$. The higher the value of $S$, the more similar two images are. This means that two images can be considered as visually similar if the correlation coefficient of their hashes is higher than a given threshold $T$. Otherwise, they will be considered as different images. 

\par

To analyze the robustness property of four hashing methods, we generated visually identical versions of 35 images from USC-SIPI \cite{USC-SIPI} dataset, we performed the following content-preserving operations, i.e. attacks: brightness adjustment, gamma correction, $3\times3$ Gaussian low-pass filtering, multiplicative noise, salt and pepper noise, JPEG compression, rotation, scaling and watermark embedding. We generated 88 new similar \textit{attacked} versions per image of the dataset. All the attacks were performed automatically using Python 3. Robustness means that visually identical images should have similar or very identical hash codes wherever their digital representations are the same or not. More particularly, the robustness of a perceptual hashing approach, an intra-test is performed, which consists of comparing, for each of the content-preserving operations, the hash codes of original images with their corresponding attacked versions using the correlation coefficient (Equation \eqref{eq:corrCoeff}). This process has been repeated for each of the content preserving operations.

Table \ref{tab:USCMeanScore} presents the perceptual robustness of four traditional state-of-the-art hashing methods, i.e.,RP-IVD \cite{Qin16PerImaHashSalientStrucFeat}, SS-Salient-SF \cite{Tang1_16RobImaHashRingPartition}, pHash \cite{Zauner10ImplemPercImaHashFunc}, and F-DNS \cite{Rubel2020F-DNS} on the USC-SIPI dataset using the average Correlation Coefficient scores. The USC-SIPI image database is a well-known image dataset that comprises various sizes of images such as $256\times256$ pixels, $512\times512$ pixels, or $1024\times1024$ pixels. This dataset is mainly used for verifying the robustness of image hashing methods.

\begin{table}[!ht]
\caption{Mean Correlation Coefficient scores analysis of four different perceptual hashing methods using USC-SIPI \cite{USC-SIPI} dataset. 35 images were used to prepare duplicates using ten different content preserving operations.}
\begin{center}
\resizebox{\textwidth}{!}{
\begin{tabular}{c@{\hspace{0.8cm}}c@{\hspace{0.5cm}}c@{\hspace{0.5cm}}c@{\hspace{0.5cm}}c@{\hspace{0.5cm}}c}
\hline
\multirow{ 2}{*}{\textbf{Operation}} & \textbf{SS-Salient-SF} & \textbf{RP-IVD} & \textbf{pHash} & \textbf{F-DNS}\\
 & \textbf{\cite{Qin16PerImaHashSalientStrucFeat}} & \textbf{\cite{Tang1_16RobImaHashRingPartition}} & \textbf{\cite{Zauner10ImplemPercImaHashFunc}} & \textbf{(Ours)} & \\
\hline
Brightness adjustment & 0.9448 & 0.9583 & 0.9884 & \textbf{0.9985}  \\
\hline
Contrast adjustment & 0.8942 & 0.9920 & 0.9967 & \textbf{0.9993}   \\
\hline
Gamma correction & 0.9719 & 0.9957 &0.9990 & \textbf{0.9995}   \\
\hline
Salt and pepper noise & 0.9963 & 0.9872 & 0.9612 & \textbf{0.9999}  \\
\hline
Multiplicative noise & 0.9947 & 0.9939 & 0.9754 & \textbf{0.9999}   \\
\hline
$3 \times 3 $ Gaussian & \multirow{ 2}{*} {0.9927} & \multirow{ 2}{*} {0.9973} & \multirow{ 2}{*} {0.9988} & \multirow{ 2}{*} {\textbf{0.9999}} \\
low-pass filter & & \\
\hline
JPEG compression & \textbf{0.9997} & 0.9986 & 0.9979 & 0.9993 \\
\hline
Scaling & 0.9723 & 0.9773 & 0.9704 & \textbf{0.9875} \\
\hline
Rotation & 0.0438 & 0.2959 & 0.2773 & \textbf{0.9365}  \\
\hline
Watermark embedding & \textbf{0.9989} & 0.9601 & 0.9894 & \textbf{0.9989} \\
\hline
\end{tabular}
}
\end{center}
\label{tab:USCMeanScore}
\end{table}

Table \ref{tab:USCMeanScore} shows the results of the average correlation coefficient score of each attack. It can be observed that F-DNS achieves the highest mean correlation coefficient scores for all the content-preserving operations, except for JPEG compression and Watermark embedding. 

We noticed that F-DNS achieved a remarkable performance in the case of the rotation, with a correlation coefficient of $0.9365$, while the best of the other methods, RP-IVD, obtained $0.2959$. Particularly, the F-DNS image hashing scheme achieves better performances of perceptual robustness, especially in rotation than the other three schemes.

In conclusion, the methods described in this document are intended for image authentication or tamper detection or image indexing. Besides, the hash can be applied to distinguish the forged, similar, and different images. Finally, we also found that a perceptual hashing algorithm should be sensitive to content-preserving operations.

In the future, a better optimization method to yield low dimensional
hash code with shorter hash code lengths is required while keeping better robustness and discrimination capabilities.

\section*{Acknowledgment}

This work was supported by the framework agreement between the University of Le\'on and INCIBE (Spanish National Cybersecurity Institute) under Addendum 01.

\medskip

\bibliography{references.bib} 

\newpage




\end{document}